\pdfoutput=1

\documentclass[11pt]{article}

\usepackage[preprint]{acl}

\usepackage{times}
\usepackage{latexsym}

\usepackage[T1]{fontenc}

\usepackage[utf8]{inputenc}

\usepackage{microtype}

\usepackage{inconsolata}

\usepackage{graphicx}
\usepackage[utf8]{inputenc} 
\usepackage[T1]{fontenc}    
\usepackage{hyperref}       
\usepackage{url}            
\usepackage{booktabs}       
\usepackage{amsfonts}       
\usepackage{nicefrac}       
\usepackage{microtype}      
\usepackage{xcolor}         
\usepackage[utf8]{inputenc}
\usepackage[T1]{fontenc}
\usepackage{bm}
\usepackage{type1cm}
\usepackage{lettrine}
\usepackage{amsmath,amssymb,amsthm}
\usepackage{moreverb}
\usepackage{mathtools}
\usepackage{array}
\usepackage{amsmath}
\usepackage{amssymb}
\usepackage{algorithm}
\usepackage{algorithmic}
\usepackage{graphics}
\usepackage{caption}
\usepackage{extarrows}
\usepackage{color}
\usepackage{framed}
\usepackage{wrapfig}
\usepackage{bm}
\usepackage{mathrsfs}
\usepackage{mathabx}
\usepackage{multirow}
\usepackage{longtable}
\usepackage{hyperref}
\usepackage{paralist}
\usepackage{indentfirst}
\usepackage{relsize}
\usepackage{extarrows}
\usepackage{lineno}
\usepackage{upgreek}
\usepackage{bm}
\usepackage{mwe}
\usepackage{booktabs}
\usepackage{authblk}
\usepackage{subcaption}
\usepackage{lettrine}
\usepackage{type1cm}
\usepackage{threeparttable}
\usepackage[figurename=Figure]{caption}
\usepackage[normalem]{ulem}
\usepackage{tcolorbox}
\tcbuselibrary{breakable}
\usepackage{float}
\usepackage{enumitem}
\title{CharacterBox: Evaluating the Role-Playing Capabilities of LLMs in Text-Based Virtual Worlds}

\author{
    \textbf{Lei Wang\textsuperscript{1}},
    \textbf{Jianxun Lian\textsuperscript{2}},
    \textbf{Yi Huang\textsuperscript{1}},
    \textbf{Yanqi Dai\textsuperscript{1}},
    \textbf{Haoxuan Li\textsuperscript{3}},
\\
    \textbf{Xu Chen\textsuperscript{1}\thanks{Corresponding Author: xu.chen@ruc.edu.cn}},
    \textbf{Xing Xie\textsuperscript{2}},
    \textbf{Ji-Rong Wen\textsuperscript{1}}
\\
\textsuperscript{1}Renmin University of China,
\textsuperscript{2}Microsoft Research Asia,
\textsuperscript{3}Peking University
\\
\texttt{\{wanglei154,xu.chen\}@ruc.edu.cn}
}

\begin{document}
\maketitle
\begin{abstract}
Role-playing is a crucial capability of Large Language Models (LLMs), enabling a wide range of practical applications, including intelligent non-player characters, digital twins, and emotional companions. Evaluating this capability in LLMs is challenging due to the complex dynamics involved in role-playing, such as maintaining character fidelity throughout a storyline and navigating open-ended narratives without a definitive ground truth. Current evaluation methods, which primarily focus on question-answering or conversational snapshots, fall short of adequately capturing the nuanced character traits and behaviors essential for authentic role-playing.
In this paper, we propose CharacterBox, which is a simulation sandbox designed to generate situational fine-grained character behavior trajectories. These behavior trajectories enable a more comprehensive and in-depth evaluation of role-playing capabilities. CharacterBox consists of two main components: the character agent and the narrator agent. The character agent, grounded in psychological and behavioral science, exhibits human-like behaviors, while the narrator agent coordinates interactions between character agents and environmental changes. Additionally, we introduce two trajectory-based methods that leverage CharacterBox to enhance LLM performance. To reduce costs and facilitate the adoption of CharacterBox by public communities, we fine-tune two smaller models, CharacterNR and CharacterRM, as substitutes for GPT API calls, and demonstrate their competitive performance compared to advanced GPT APIs. The code is available at \url{https://github.com/Paitesanshi/CharacterBox}.
\end{abstract}

\section{Introduction}

Role-playing is an advanced capability of large language models (LLMs) that allows them to mimic human-like behavior within the context of specific roles. This functionality underpins various practical applications, such as intelligent non-player characters (NPCs) in video games, digital replicas for personal assistants, and emotional support in mental healthcare. While there are comprehensive benchmarks for evaluating the general-purpose abilities of LLMs, including language understanding~\cite{hendryckstest2021}, conversation~\cite{chiang2024chatbot}, and reasoning~\cite{Clark2018ThinkYH}, the assessment of role-playing capabilities remains an area that is not as thoroughly explored. 
Current evaluation methods, such as static self-reporting questionnaires~\cite{jiang2024evaluating} and simple dialogue tasks~\cite{tu2024charactereval}, fail to capture the full complexity of role-specific behaviors in real-life scenarios.
These methods are limited by their static nature and inability to reflect continuous role-playing interactions~\cite{ahn2024timechara}. In reality, a character’s actions, attitudes, and emotions are dynamic and evolve in response to the surrounding environment and other individuals. The proverb "A man is judged by his deeds, not by his words" applies here: an LLM’s true role-playing ability cannot be fully understood from static dialogues or self-reports alone, but rather is demonstrated during interaction with its environment~\cite{chen2024oscars}.

\begin{figure*}[t!]
    \centering
    \includegraphics[width=\linewidth]{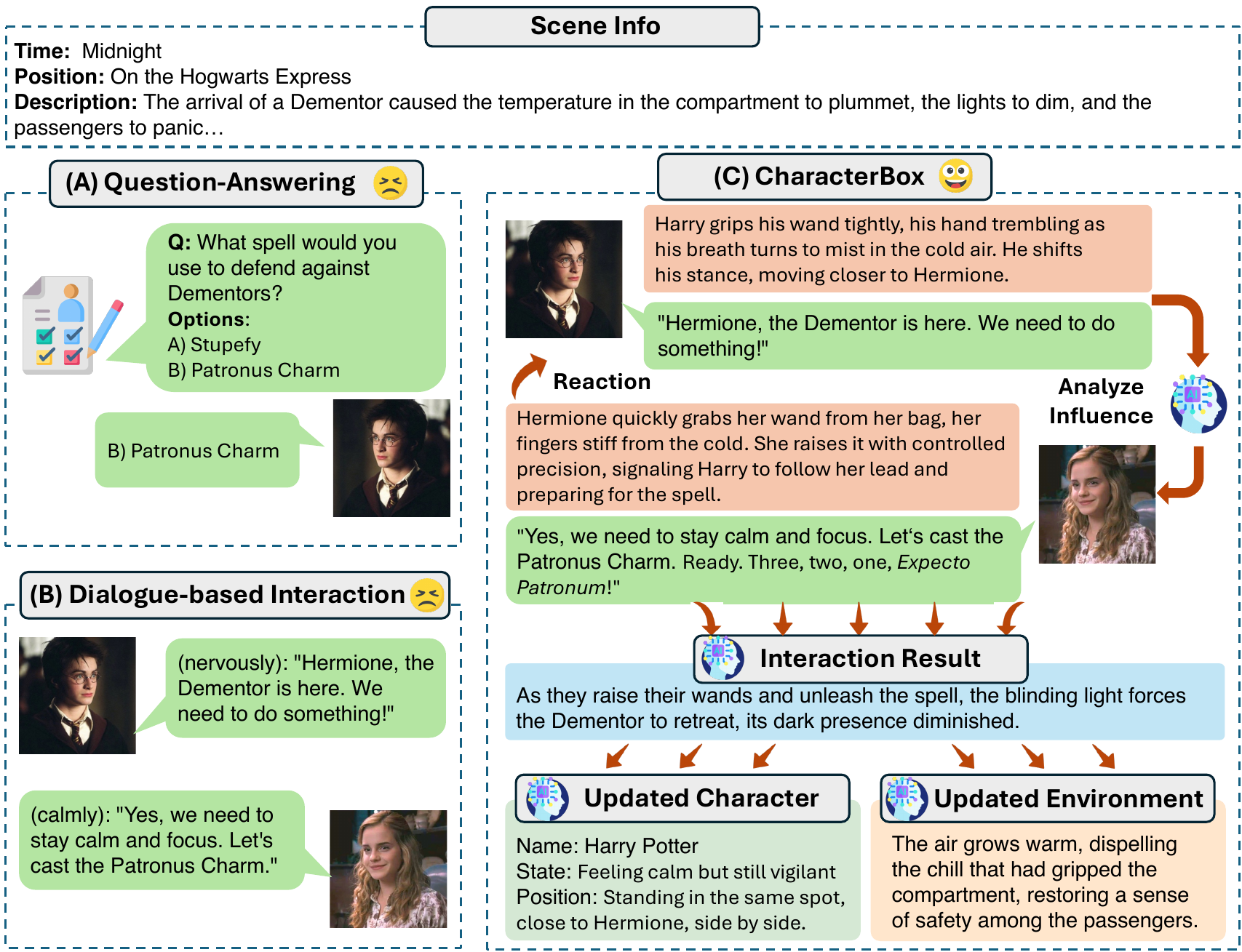}
    \caption{A comparison of different role-playing facilities: (A) self-reported QA; (B) Conversations; and (C) CharacterBox. Unlike the other methods, CharacterBox not only prompts role agents for utterances and actions but also includes components to track environmental changes and coordinate interactions between role agents.}
    \label{comp}
\end{figure*}

In this paper, we present CharacterBox, a  dynamic, multi-agent virtual world tailor-made for eliciting nuanced human-like behaviors from LLMs in the context of role-playing evaluations. CharacterBox crafts immersive scenarios tailored to specific roles, incorporating detailed role specifications, contextual backgrounds, and interactions that mirror real-world complexity.  LLMs are assigned roles and interact with both the environment and other characters through dialogue and actions that reflect their role-specific traits. A comparison between previous methods and CharacterBox is presented in Fig~\ref{comp}.

To track the evolving states of both characters and their surroundings, we incorporate a narrator component, typically powered by advanced models like GPT-3.5-turbo. The narrator monitors character actions and environmental changes, generating behavior trajectories used to assess LLM role-playing performance. 

Given the subjective nature of evaluating behavior trajectories, we further employ GPT-4 as a reward model to assess role-playing performance from seven distinct perspectives. This approach enables us to compare different LLMs based on their interactive role-play abilities. To reduce the dependency on costly APIs, we fine-tune two smaller language models, named CharacterNR and CharacterRM, to function as the narrator and reward model by distilling knowledge from the superior teacher models, GPT-3.5 and GPT-4, respectively. This allows our evaluation pipeline to operate independently, free from API costs.
 
Our benchmark reveals notable discrepancies in role-playing abilities between LLMs. 
Furthermore, we introduce guided and reflective trajectory fine-tuning. The guided method uses high-quality behavior trajectories to shape model behavior, while the reflective method allows models to self-correct based on their own generated trajectories. Both methods significantly improve role-playing performance across evaluation dimensions.

In summary, the key contributions are:

$\bullet$~We introduce CharacterBox, the first dynamic, multi-agent interactive virtual world tailored for role-playing evaluations. The framework features character agents built on well-structured modules, along with narrator agent that dynamically updates both the characters and environment, creating realistic, evolving interactions.

$\bullet$~We construct a comprehensive benchmark to evaluate role-playing capabilities of LLMs, testing a wide range of models, both closed-source and open-source. Our experiments validate the reliability and validity of this benchmark.

$\bullet$~We introduce two trajectory-based fine-tuning methods—guided and reflective—that significantly enhance LLMs' role-playing abilities. By leveraging behavior trajectories generated by CharacterBox, smaller models such as 7B LLMs achieve performance levels comparable to advanced models like GPT-3.5-turbo.
 
$\bullet$~We fine-tune two essential components, CharacterNR and CharacterRM, to create a cost-efficient, self-contained pipeline, significantly reducing dependency on expensive GPT API calls while maintaining high-quality role-playing performance assessments.

\section{Related Work}

\subsection{Evaluation of Role-Playing Agent}

Evaluating the role-playing capabilities of LLMs is essential yet challenging, leading researchers to propose various benchmarks~\cite{xu2024character,yuan2024evaluating,shao2023character}. RoleBench~\cite{wang2023rolellm} offers a role-granular dataset with extensive role dialogues for evaluation. CharacterEval~\cite{tu2024charactereval} uses dialogues from 77 characters in Chinese scripts, with 14 evaluation metrics and a reward model. InCharacter~\cite{wang2023incharacter} tests role fidelity by converting psychological scales into interview formats. RoleInteract~\cite{chen2024roleinteract} evaluates RPAs in individual and group interactions, assessing social behaviors based on the roles. However, these benchmarks focus on static dialogues or QA interactions, while CharacterBox expands the evaluation to dynamic scenarios, including specific actions.

\subsection{LLM-based virtual environment}
Based on extensive training data, LLMs possess logical reasoning abilities and vast knowledge, making it possible to construct virtual environments based on LLMs~\cite{zhang2023generative, williams2023epidemic}. GenerativeAgent~\cite{park2023generative} manually designs a virtual town, allowing LLM-based agents to play different roles to simulate human life in the town and interact with other agents. RecAgent~\cite{wang2023recagent} built a virtual recommendation platform, where agents as users can browse recommended movies and chat and post on social platforms. UGI~\cite{xu2023urban} constructed a city simulation platform based on the real world, where agents can engage in social interactions, street navigation, and other urban behaviors. However, these LLM-based virtual environments are time-consuming to meticulously design and pre-define, cannot be dynamically updated, and are difficult to create in large quantities. Our framework, CharacterBox, can dynamically update the environment according to the agents' behaviors within it and can extract or create new scenes based on a given context.

\section{Evaluation Framework based on Text-based Interactive Virtual World}

In this section, we present an in-depth exploration of the interactive evaluation framework, CharacterBox. The CharacterBox workflow is structured around three pivotal phases: scene crafting, autonomous story play, and evaluation. 

\subsection{Scene Crafting}\label{scene_generation}
Scenes form the foundation of our evaluation framework. A scene, represented as $S$, includes environmental and character elements. Environmental aspects cover time, location, and descriptions that influence character behavior. Character information includes profiles like names, roles, physical and psychological states. Formally, a scene with $n$ characters is: $S = \{E, C\}$, where $E$ is the environment and $C = \{c_1, c_2, \dots, c_n\}$ are the characters.

When LLMs engage in role-playing using scenes drawn from novels or scripts, there is a risk of replicating content already present in their training data~\cite{li2024task}. To address this, the generation of original scenes becomes necessary, but also more challenging.
To ensure high-quality scene creation, we divided the development process into three stages, assigning LLMs the roles of screenwriter, director, and evaluator~\cite{li2024camel,qian2023communicative}. As screenwriters, LLMs extract or generate scenes that align with the story's logic. As directors, they refine these scenes by focusing on key elements like events, character details, and interactions to maintain coherence and engagement. Finally, as evaluators, LLMs assess the scenes based on creativity, coherence, conformity, and detail, accepting only those that meet quality standards. These curated scenes initiate CharacterBox, providing a dynamic stage for interactive role play.

\subsection{Autonomous Story Play}

Following the scene crafting phase, the environment 
$E$ serves as the stage and the characters $C$ as the actors in the autonomous story play. Moreover, we design the narrator $NR$ as a world model to analyze the characters' actions and update both the environment and character states in real time. In this way, the scene transforms from a static setting into a dynamic virtual world that evolves as the story progresses.

\textbf{Environment}. The environment includes time, location, and descriptions, which are dynamically influenced by character actions. The narrator updates these elements in real-time.

\textbf{Character}. Characters, controlled by LLMs, use a memory module inspired by~\cite{park2023generative}, where each agent utilizes a vector database to record past actions and observations, retrieving relevant information to guide future behavior. Each character maintains self-beliefs and environment-beliefs following the Belief-Desire-Intention (BDI) model~\cite{georgeff1999belief}. Self-beliefs include identity, self-awareness, and goals, while environment-beliefs represent the character's understanding of the surroundings and other agents.

During story play, characters take turns planning and executing their actions at the start of each round, drawing on memory and the BDI model, as inspired by prior work~\cite{peinado2008revisiting}. Actions are expressed in detailed descriptions, and characters can respond immediately to others. After each round, both self-beliefs and environment-beliefs are updated accordingly.

\textbf{Narrator}. The narrator serves as an objective world model, responsible for accurately analyzing the development of characters and the environment within CharacterBox. As the core of the framework, the narrator performs the following functions:

$\bullet$ \textbf{Analyze Action Influence}: When a character \(c_i\) takes an action, the narrator assesses its impact on other characters by considering their current states. The narrator identifies the character \(c_r\) most affected and likely to respond to \(c_i\). The action \(a_i\) and resulting influence $f_r$ are conveyed to \(c_r\).  

$\bullet$ \textbf{Analyze Interaction Result}: The narrator determines the outcome of the interaction between \(c_i\) and \(c_r\), represented by \(R\). This outcome is used to update both characters’ memories, physical positions, and psychological states.  

$\bullet$ \textbf{Update Character}: The narrator updates each character's state based on their own action or the result of interactions. If no other character responds to \(c_i\), \(c_i\)'s state is updated based on its own action.

$\bullet$ \textbf{Update Environment}: After each round, the narrator updates the environment \(E\) based on the characters' actions and their outcomes. If no actions affect the environment, it remains unchanged.

The complete process is illustrated in Algorithm~\ref{alg:story_play}. For detailed prompts, please refer to Appendix~\ref{prompt}.
\begin{algorithm}[H]
\caption{Autonomous Story Play Process}
\label{alg:story_play}
\begin{algorithmic}[1]
\STATE \textbf{Initialize} environment $E$ and character set $C = \{c_1, c_2, \dots, c_n\}$
\WHILE{story not concluded}
    \FOR{each character $c_i \in C$}
        \STATE \textbf{Plan and perform action}:
        \STATE $a_i = \text{PlanAndPerform}(c_i, E)$
        \STATE \textbf{Narrator: Analyze influence of $a_i$}
        \STATE Determine most affected character $c_r$ and influence $f_r$: $c_r, f_r = \text{NR}(E, a_i, C)$
        \IF{$c_r$ exists}
            \STATE $c_r$ responds based on $a_i$ and $f_r$
            \STATE \textbf{Narrator: Analyze interaction result $R$} and update $c_i$, $c_r$
        \ELSE
            \STATE \textbf{Narrator: Update} $c_i$ state based on $a_i$
        \ENDIF
        \STATE \textbf{Narrator: Update environment} $E$ based on actions and interactions
    \ENDFOR
\ENDWHILE
\end{algorithmic}
\end{algorithm}

\subsection{Evaluation}\label{evaluation}

Through autonomous story play, we obtain a series of actions from each character in different contexts, forming a trajectory formally represented as $\tau=\{E,c,{o_1,a_1},{o_2,a_2},...,{o_n,a_n}\}$, where each character's actions and observations are captured in relation to the environment and character information. To comprehensively evaluate the role-playing capabilities of LLMs in long-term dynamic environments, we design metrics across three main dimensions, drawing inspiration from key aspects of effective role-playing~\cite{chen2024persona,chen2024oscars}:

\textbf{Character Fidelity} assesses how accurately the model represents the character’s knowledge and behaviors. This is crucial for maintaining consistency with the character’s identity:

$\bullet$ \textbf{Knowledge Accuracy (KA)}: Ensures information provided by character is factually correct and aligned with their background knowledge.

$\bullet$ \textbf{Behavioral Accuracy (BA)}: Measures the consistency of character’s behaviors and linguistic patterns, ensuring alignment with their traits.

\textbf{Human-Likeness} evaluates the realism and believability of the character’s portrayal, focusing on dynamic, emotionally engaging interactions:

$\bullet$ \textbf{Emotional Expression (EE)}: Evaluates the ability of the character to express emotions vividly, key to enhancing user immersion.

$\bullet$ \textbf{Personality Traits (PT)}: Determines whether the model consistently maintains the character’s core personality traits throughout interactions.

\textbf{Consistency} focuses on maintaining logical continuity in the character’s behavior across interactions, which is essential for immersive role-playing:

$\bullet$ \textbf{Immersion (IM)}: Measures the character's ability to stay in role, ensuring a continuous and believable experience for the user.

$\bullet$ \textbf{Adaptability (AD)}: Assesses how the character adjusts to evolving situations while maintaining their integrity.

$\bullet$ \textbf{Behavioral Coherence (BC)}: Evaluates the logical consistency of character’s actions in relation to previous behaviors and current context.

Each metric is scored from 1 to 5, with higher scores indicating stronger performance. To enhance evaluation accuracy, we leverage GPT-4 to first generate a critique of the character’s trajectory, integrating this critique into the prompt before assessing each criterion. These metrics collectively ensure that the role-playing agents are not only accurate and engaging but also capable of sustaining character fidelity over extended interactions, which is crucial for immersive narrative experiences.

\section{Enhancing Role-playing Ability with Trajectories}

CharacterBox facilitates the efficient generation of character trajectories across diverse scenes, providing valuable insights into character reactions and behaviors within varied contexts. These trajectories offer a unique opportunity to enhance the role-playing capabilities of language models. To leverage this data, we propose two distinct methods for fine-tuning LLMs using generated trajectories:

\textbf{Guided Trajectory Fine-tuning}. We first assess the role-playing capabilities of LLMs using CharacterBox, selecting high-performing models as teachers. The trajectories generated by these models are then used to fine-tune student models, resulting in significant improvements in the latter’s ability to simulate complex character interactions.

\textbf{Reflective Trajectory Fine-tuning}. In this approach, we explore the self-reflective capabilities of LLMs. Models analyze their own generated trajectories, identifying inconsistencies and areas for improvement in character portrayal. The models then rewrite these trajectories to enhance character consistency and depth. These revised trajectories are subsequently used for fine-tuning, further strengthening the model's capacity to simulate realistic and nuanced interactions.

\section{Building for a Self-contained Evaluation Workflow}

In CharacterBox, the narrator agent and evaluation agent can be powered by advanced language models like GPT-4 or individuals familiar with the characters. However, these methods are costly and lack scalability. To address this, we develop CharacterNR and CharacterRM to reduce costs and enhance scalability.

\textbf{CharacterNR}.
\label{sec:characterNR}
CharacterNR acts as the narrator within CharacterBox. Initially, GPT-3.5 is used to generate narrator trajectory data due to its strong instruction-following abilities. To handle both Chinese and English scenes, we select Qwen2.5-7B as the base model and fine-tune it using LoRA~\cite{hu2021lora} with data generated by GPT-3.5.

\textbf{CharacterRM}.
\label{sec:characterRM}
We collect evaluation scores from GPT-4 across 100 scenes, incorporating outputs from nine different LLMs to ensure diversity. To maintain fairness in scoring, we select ChatGLM3-6B~\cite{glm2024chatglm} as the base model, since it is not among the evaluated models. We then fine-tune it using LoRA on the collected data, resulting in CharacterRM.

\section{Experiment}

\subsection{Evaluation Setting}
$\bullet$ \textbf{Scene}. We select 10 well-known novels and scripts as scene sources, covering a range of settings and themes (see Appendix~\ref{source} for details). Five works are in Chinese and five in English, with evaluations conducted in both language settings. Each scene includes specific environment and character information, with 2 to 4 characters per scene (see Appendix~\ref{scene_statistics} for further details).

$\bullet$ \textbf{LLM}. We evaluate the role-playing ability of nine LLMs varying in model size. For closed-source models, we use \texttt{GPT-4-Turbo-1106-preview} as GPT-4~\cite{achiam2023gpt} and \texttt{GPT-3.5-Turbo-1106} as GPT-3.5~\cite{brown2020language}. For open-source models, we evaluate Baichuan2-7B/13B~\cite{yang2023baichuan}, Qwen2.5-7B/14B~\cite{bai2023qwen}, Mistral-7B-v0.2~\cite{jiang2023mistral}, Llama3-8B~\cite{touvron2023llama}, and Phi-3.5-mini~\cite{abdin2024phi}. All open-source LLMs we evaluated are versions that have undergone instruction tuning.

\subsection{Overall Performance}

We select five existing and five new scenes for each novel or script, resulting in 50 English and 50 Chinese scenes. Each LLM’s performance is assessed by evaluating the behavior trajectories of characters in each scene, with the average score representing the LLM’s performance for that scene. The overall score for each LLM is then calculated by averaging scores across all 50 scenes.

\begin{table*}[!ht]
    \centering
    
    \renewcommand\arraystretch{1.1}
    \begin{threeparttable}
        \scalebox{0.9}{
    \begin{tabular}{
    p{2.5cm}<{\centering}p{1.25cm}<{\centering}
    p{1.25cm}<{\centering}p{1.25cm}<{\centering}
    p{1.25cm}<{\centering}p{1.25cm}<{\centering}
    p{1.25cm}<{\centering}p{1.3cm}<{\centering}|p{1.5cm}<{\centering}}
    \hline
        Model & KA & BA & EE & PT & IM & AD & BC & Average \\ \hline
         \multicolumn{9}{c}{\textbf{English Scene}} \\ \cline{1-9}
Phi-3.5-mini & 3.014$_{\pm .55}$ & 2.521$_{\pm .48}$ & 2.775$_{\pm .53}$ & 2.676$_{\pm .53}$ & 2.535$_{\pm .54}$ & 2.437$_{\pm .51}$ & 2.620$_{\pm .54}$ & 2.654$_{\pm .48}$ \\
Mistral-7B-v0.2 & 2.525$_{\pm .57}$ & 2.406$_{\pm .48}$ & 3.099$_{\pm .53}$ & 2.891$_{\pm .53}$ & 2.960$_{\pm .54}$ & 3.050$_{\pm .51}$ & 2.802$_{\pm .54}$ & 2.819$_{\pm .48}$ \\
Baichuan2-7B & 3.041$_{\pm .51}$ & 2.786$_{\pm .44}$ & 2.602$_{\pm .51}$ & 3.041$_{\pm .48}$ & 2.857$_{\pm .46}$ & 2.592$_{\pm .46}$ & 2.969$_{\pm .50}$ & 2.841$_{\pm .44}$ \\
Llama-3-8B & 3.191$_{\pm .59}$ & 2.882$_{\pm .54}$ & 2.836$_{\pm .49}$ & 3.245$_{\pm .53}$ & 3.091$_{\pm .54}$ & 2.573$_{\pm .51}$ & 3.109$_{\pm .54}$ & 2.990$_{\pm .48}$ \\
Baichuan2-13B & 3.237$_{\pm .49}$ & 3.062$_{\pm .45}$ & 2.959$_{\pm .37}$ & 3.289$_{\pm .46}$ & 3.186$_{\pm .42}$ & 3.082$_{\pm .40}$ & 3.247$_{\pm .46}$ & 3.152$_{\pm .39}$ \\
Qwen2.5-7B & 2.202$_{\pm .55}$ & \underline{3.753}$_{\pm .48}$ & \underline{3.400}$_{\pm .53}$ & 3.653$_{\pm .53}$ & 3.030$_{\pm .54}$ & \underline{3.374}$_{\pm .51}$ & 3.644$_{\pm .54}$ & 3.294$_{\pm .48}$ \\
Qwen2.5-14B & 3.130$_{\pm .56}$ & \textbf{3.967}$_{\pm .55}$ & 2.900$_{\pm .45}$ & \underline{3.860}$_{\pm .49}$ & 3.574$_{\pm .48}$ & 3.016$_{\pm .43}$ & \underline{3.984}$_{\pm .51}$ & 3.490$_{\pm .45}$ \\
GPT-3.5 & \underline{3.702}$_{\pm .57}$ & 3.681$_{\pm .52}$ & 3.186$_{\pm .40}$ & \underline{3.867}$_{\pm .44}$ & \underline{3.717}$_{\pm .40}$ & 3.159$_{\pm .44}$ & 3.841$_{\pm .49}$ & \underline{3.593}$_{\pm .42}$ \\
GPT-4 & \textbf{3.796}$_{\pm .49}$ & 3.746$_{\pm .45}$ & \textbf{3.789}$_{\pm .39}$ & \textbf{3.974}$_{\pm .36}$ & \textbf{4.088}$_{\pm .33}$ & \textbf{3.930}$_{\pm .44}$ & \textbf{4.158}$_{\pm .35}$ & \textbf{3.926}$_{\pm .36}$ \\ \hline
\multicolumn{9}{c}{\textbf{Chinese Scene}} \\ \hline
Phi-3.5-mini & 2.800$_{\pm .55}$ & 2.554$_{\pm .43}$ & 2.662$_{\pm .57}$ & 2.615$_{\pm .50}$ & 2.539$_{\pm .52}$ & 2.585$_{\pm .45}$ & 2.585$_{\pm .51}$ & 2.620$_{\pm .50}$ \\
Mistral-7B-v0.2 & 2.878$_{\pm .59}$ & 2.791$_{\pm .39}$ & 2.904$_{\pm .60}$ & 3.000$_{\pm .56}$ & 3.035$_{\pm .56}$ & 2.939$_{\pm .38}$ & 2.922$_{\pm .58}$ & 2.924$_{\pm .52}$ \\
Llama-3-8B & 3.452$_{\pm .49}$ & 3.278$_{\pm .36}$ & 2.730$_{\pm .49}$ & 3.426$_{\pm .50}$ & 3.209$_{\pm .45}$ & 2.870$_{\pm .35}$ & 3.435$_{\pm .49}$ & 3.200$_{\pm .45}$ \\
Baichuan2-7B & 3.763$_{\pm .43}$ & 3.535$_{\pm .40}$ & 3.123$_{\pm .56}$ & 3.728$_{\pm .54}$ & 3.570$_{\pm .42}$ & 3.149$_{\pm .38}$ & 3.640$_{\pm .54}$ & 3.501$_{\pm .47}$ \\
Baichuan2-13B & 3.617$_{\pm .40}$ & 3.522$_{\pm .49}$ & 3.270$_{\pm .49}$ & 3.713$_{\pm .50}$ & 3.557$_{\pm .44}$ & 3.243$_{\pm .42}$ & 3.635$_{\pm .52}$ & 3.508$_{\pm .46}$ \\
GPT-3.5 & 3.861$_{\pm .45}$ & 3.783$_{\pm .34}$ & 3.243$_{\pm .42}$ & 4.000$_{\pm .43}$ & 3.774$_{\pm .33}$ & 3.313$_{\pm .33}$ & 3.904$_{\pm .41}$ & 3.697$_{\pm .39}$ \\
Qwen2.5-7B & \textbf{4.341}$_{\pm .50}$ & 3.951$_{\pm .39}$ & 3.289$_{\pm .32}$ & 4.026$_{\pm .39}$ & 3.871$_{\pm .33}$ & 3.196$_{\pm .29}$ & 3.982$_{\pm .33}$ & 3.808$_{\pm .37}$ \\
Qwen2.5-14B & 4.057$_{\pm .42}$ & \underline{4.122}$_{\pm .31}$ & \underline{3.743}$_{\pm .39}$ & \underline{4.321}$_{\pm .39}$ & \underline{4.042}$_{\pm .30}$ & \underline{3.742}$_{\pm .27}$ & \underline{4.369}$_{\pm .34}$ & \underline{4.057}$_{\pm .35}$ \\
GPT-4 & \underline{4.252}$_{\pm .45}$ & \textbf{4.357}$_{\pm .39}$ & \textbf{4.096}$_{\pm .30}$ & \textbf{4.496}$_{\pm .33}$ & \textbf{4.530}$_{\pm .30}$ & \textbf{4.139}$_{\pm .24}$ & \textbf{4.522}$_{\pm .34}$ & \textbf{4.342}$_{\pm .34}$ \\
        \hline
    \end{tabular}
    }
    \end{threeparttable}
    \caption{Evaluation results on English and Chinese scenes. Each value is presented as mean$_{\pm \text{standard deviation}}$. \textbf{Bold} values indicate the highest scores, and \underline{underlined} values indicate the second-highest scores.}
    \label{overall}
\end{table*}

Table \ref{overall} presents the results across seven metrics for both English and Chinese scenes. GPT-4 performs the best across both English and Chinese scenes. GPT-3.5 shows strong performance in English scenes but falls behind Qwen2.5 models, especially Qwen2.5-14B, in Chinese scenes. The latter surpasses GPT-3.5 in multiple metrics and approaches GPT-4’s competitiveness. Qwen2.5 and Baichuan2 models, due to their large-scale training on Chinese corpora, demonstrate a clear advantage in Chinese scenarios. In contrast, models like Mistral-7B-v0.2 and Llama3-8B perform better in English scenes but are relatively weaker in Chinese. Overall, bilingual models, especially Qwen2.5 and Baichuan2, show stronger role-playing capabilities in Chinese scenes, highlighting the impact of language-specific training on role-playing abilities.

\subsection{Reliability and Validity of CharacterBox}\label{reliability_validity}

\textbf{Reliability}. We measure reliability of CharacterBox using Cronbach's alpha to assess internal consistency~\cite{cronbach1951coefficient}, following prior works~\cite{yang2024llm}. As shown in Table~\ref{table:reliability}, CharacterBox achieves high Cronbach's alpha values across three evaluation dimensions in both English and Chinese scenes. The consistently high scores, with most above 0.9, indicate that CharacterBox provides a reliable evaluation of LLMs' role-playing capabilities across different scenarios.


\begin{table}[H]
    \centering
    
    \renewcommand\arraystretch{0.8}
    \begin{tabular}{
        p{4cm}<{\centering}p{1cm}<{\centering}p{1cm}<{\centering}}
        \toprule
        Cronbach alpha & English & Chinese \\
        \midrule
        Character Fidelity & 0.958 & 0.951 \\
        Human-Likeness & 0.832 & 0.862 \\
        Consistency & 0.945 & 0.941 \\
        \bottomrule
    \end{tabular}
    \caption{Cronbach alpha values of CharacterBox across three evaluation dimensions.}
    \label{table:reliability}
\end{table}



\begin{table*}[h]
\centering
    \renewcommand\arraystretch{0.9}
    \begin{threeparttable}
        \scalebox{1}{
        \begin{tabular}{ccccccccc}
        \hline
        Model & KA & BA & EE & PT & IM & AD & BC & Overall \\ \hline
        GPT-4 & 0.445 & 0.475 & \textbf{0.597} & 0.445 & 0.618 & \textbf{0.742} & \textbf{0.601} & \textbf{0.688}\\ 
        ChatGLM & 0.422 & 0.334 & 0.407 & 0.151 & 0.497 & 0.386 & 0.321 & 0.482 \\ 
        CharacterRM & \textbf{0.681} & \textbf{0.584} & 0.464 & \textbf{0.464} & \textbf{0.620} & 0.434 & 0.567 & 0.610 \\ \hline
        \end{tabular}
        }
\end{threeparttable}
\caption{Pearson correlation coefficient between GPT-4, ChatGLM, CharacterRM, and human expert evaluation results. \textbf{Bold} values highlight the highest correlation for each metric.}
\label{tab:story_rm}
\end{table*}

\textbf{Validity}. To validate our evaluation, we enlist three experts familiar with both the five Chinese and five English scenes used in the assessment to rate the character trajectories. We calculate the Pearson correlation coefficient between CharacterBox scores and expert ratings, using GPT-4 as the evaluator. The strong correlation of 0.688, as shown in Table \ref{tab:story_rm}, confirms that CharacterBox’s automated evaluations closely align with human assessments. This consistency underscores CharacterBox's effectiveness in evaluating LLMs' role-playing capabilities. Additionally, Table \ref{overall} shows that larger models, such as Qwen2.5-14B versus Qwen2.5-7B and Baichuan2-13B versus Baichuan2-7B, consistently outperform their smaller versions, reinforcing the common belief that model size correlates with improved performance.


\begin{figure*}[ht]
    \centering
    \includegraphics[width=\linewidth]{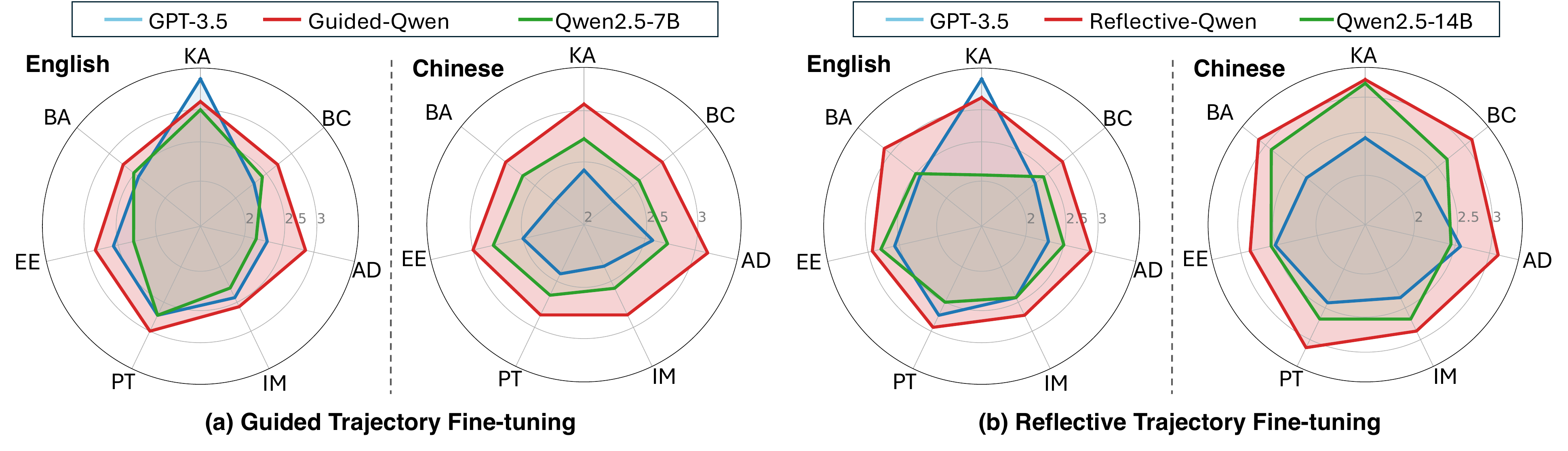}
    \caption{Performance comparison under Guided and Reflective Trajectory Fine-tuning across English and Chinese scenes.}
    \label{te-llm}
\end{figure*}
\subsection{Role-playing Ability of Trajectory Enhanced LLM}

We fine-tune Qwen2.5-7B and Qwen2.5-14B models using LoRA, applying two strategies: Guided and Reflective Trajectory fine-tuning. The performance of the fine-tuned models is evaluated on five newly generated English scenes and five Chinese scenes, which were not part of the training data.

\textbf{Guided Trajectory Fine-tuning}.
In this method, Qwen2.5-7B is fine-tuned with high-quality trajectories from CharacterBox. These trajectories are selected from the top-performing models across both languages in Table~\ref{overall}. As shown in Fig~\ref{te-llm}(a), Guided-Qwen improves by 14.3\% overall in English scenes and 10.7\% in Chinese scenes.  In some categories such as EE and AD, the Guided-LLM outperforms GPT-3.5, demonstrating the effectiveness of using high-quality trajectories to enhance LLM's role-playing capability.

\textbf{Reflective Trajectory Fine-tuning}.
For the reflective approach, we utilize Qwen2.5-14B, leveraging its capacity to handle the complexity of iterative improvements. The model is fine-tuned with rewritten trajectories, allowing it to reflect on its initial outputs and generate refined responses. As illustrated in Fig~\ref{te-llm}(b), Reflective-Qwen improves by 19.9\% in English scenes and 12.8\% in Chinese scenes, outperforming the base model across all metrics. Notably, Reflective-Qwen also achieves greater gains than Guided-Qwen, suggesting that the reflective process enables the model to generate more contextually nuanced and refined responses, leading to more believable role-playing performance.

These findings demonstrate that role-playing abilities in LLMs can be significantly enhanced by learning from well-constructed trajectories. The guided trajectory fine-tuning method provides the model with diverse, detailed character responses, while reflective fine-tuning encourages the model to iteratively improve its own outputs. By integrating these strategies, we show that CharacterBox can effectively generate character trajectories that lead to substantial improvements in role-playing performance.

\subsection{Analysis of Evaluation Stages}\label{evaluation_stages}

\begin{table*}[htbp]
    \centering
    \renewcommand\arraystretch{0.8}
     \begin{threeparttable}
        \scalebox{0.95}{
\begin{tabular}{p{2cm}<{\centering}  p{0.5cm}<{\centering} p{1.3cm}<{\centering} p{1.3cm}<{\centering}p{1.3cm}<{\centering} p{1.3cm}<{\centering} p{1.3cm}<{\centering}p{1.3cm}<{\centering}p{1.3cm}<{\centering}}
        \toprule
        \textbf{Model} & \multicolumn{2}{c}{\textbf{Creativity}} & \multicolumn{2}{c}{\textbf{Coherence}} & \multicolumn{2}{c}{\textbf{Conformity}} & \multicolumn{2}{c}{\textbf{Detail}}   \\ \cmidrule{2-3} \cmidrule{4-5}\cmidrule{6-7}\cmidrule{8-9}
        & EXT & GEN & EXT & GEN & EXT & GEN & EXT & GEN  \\
        \midrule
        GPT-4 & - & 3.1$\pm${0.35} & 3.7$\pm${0.32} & 3.6$\pm${0.26} & 3.9$\pm${0.26} & 3.4$\pm${0.42} & 3.4$\pm${0.33} & 3.6$\pm${0.40}    \\
        GPT-3.5 & - & 3.0$\pm${0.45} & 3.4$\pm${0.34} & 3.7$\pm${0.35} & 3.6$\pm${0.38} & 3.6$\pm${0.33} & 3.0$\pm${0.31} & 3.0$\pm${0.36}   \\
        ChatGLM3 & - & 3.2$\pm${0.33} & 3.4$\pm${0.45} & 3.6$\pm${0.23} & 3.4$\pm${0.33} & 4.0$\pm${0.22} & 2.7$\pm${0.45} & 3.0$\pm${0.46}   \\
        Three-Stage & - & \textbf{3.5}$\pm${0.49} & \textbf{4.0}$\pm${0.34} & \textbf{4.2}$\pm${0.21} & \textbf{4.1}$\pm${0.22} & \textbf{4.2}$\pm${0.26} & \textbf{4.1}$\pm${0.27} & \textbf{3.9}$\pm${0.26}  \\
        \bottomrule
    \end{tabular}
    }
    \end{threeparttable}
     \caption{Performance comparison of different LLMs in crafting scenes. EXT means extracting scenes. GEN means generating new scenes. The base model of Three-Stage method is ChatGLM3-6B.}
    \label{tab:scene}
\end{table*}

$\bullet$ \textbf{Three-Stage Scene Crafting}. 

While powerful models like GPT-3.5 and GPT-4 excel at scene crafting, their high cost limits large-scale use. To address this, we implement a three-stage scene crafting approach using smaller open-source models. Our method had the LLM extract and generate scenes from 10 scripts, evaluating the results from four aspects. As shown in Table~\ref{tab:scene}, GPT-4 excels at extracting scenes but has no advantage in generating new ones. In contrast, our three-stage method based on ChatGLM3-6B improves upon its baseline and outperforms GPT-3.5 and GPT-4 in both tasks. This demonstrates that small open-source LLMs can replace closed-source models in scene crafting, reducing costs significantly.

\begin{figure}[ht]
    \centering
    \includegraphics[width=\linewidth]{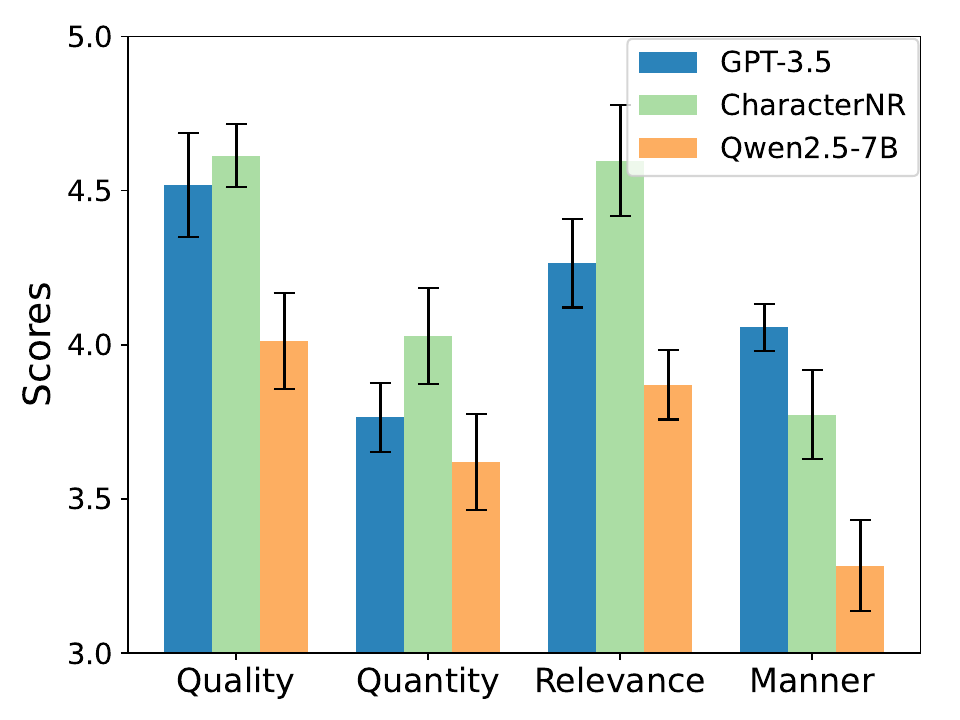}
    \caption{comparison between GPT-3.5, CharacterNR and the base model Qwen2.5-7B.}
    \label{fig:story_nr}
\end{figure}
$\bullet$ \textbf{CharacterNR}.   
To evaluate the effectiveness and generalization of our local CharacterNR, we generate five new Chinese and five new English scenes not included in the fine-tuning data. We assess the narrator's performance based on the Gricean Maxims~\cite{dale1995computational}: Quality, which reflects the accuracy and reasonableness of the results; Quantity, ensuring the information is substantial but not redundant; Relevance, which measures how pertinent the results are to the task; and Manner, assessing whether the output is vivid, expressive, and engaging, in line with CharacterBox's features.
As shown in Fig~\ref{fig:story_nr}, the fine-tuned CharacterNR significantly outperforms Qwen2.5-7B in all metrics and matches or exceeds GPT-3.5. This improvement is largely due to Qwen2.5-7B’s strong performance, especially in Chinese scenes, and its improved instruction-following ability after fine-tuning.

$\bullet$ \textbf{CharacterRM}. CharacterRM serves as the reward model for evaluating and scoring character trajectories. We select ChatGLM3-6B as the base model, fine-tuning it with GPT-4-generated evaluation results as labels. Similar to Section~\ref{reliability_validity}, we validate CharacterRM by scoring new Chinese and English scenes outside the fine-tuning data and comparing the results with human expert evaluations. As shown in Table~\ref{tab:story_rm}, CharacterRM outperforms ChatGLM3-6B in all metrics and achieves an overall correlation of 0.610, close to GPT-4’s 0.688, demonstrating its reliability and strong alignment with human evaluations.

\section{Conclusion}


In this paper, we introduce CharacterBox, a dynamic, text-based virtual environment specifically designed to evaluate the role-playing capabilities of LLMs. By creating immersive scenarios that reflect the complexities of real-world interactions, CharacterBox captures nuanced human-like behaviors in LLMs, going beyond static evaluation methods. We demonstrate that fine-tuning smaller models with high-quality behavior trajectories significantly enhances their role-playing abilities. Additionally, we develope two fine-tuned components, CharacterNR and CharacterRM, allowing for a cost-efficient and self-sustained evaluation process without relying on expensive API calls. These contributions establish CharacterBox as a powerful and self-contained tool for assessing and improving LLM role-playing performance across diverse scenarios.

\section*{Limitation} \label{limitation}

While the CharacterBox framework offers an innovative and comprehensive approach to evaluating the role-playing capabilities of LLMs, several limitations remain: First, the runtime efficiency needs to be improved to accommodate large-scale evaluation scenarios. Second, additional human-annotated data is required to better train the reward model, ensuring more accurate evaluations. Finally, the limited context window of LLMs presents a challenge in interactive role-playing, as prompts cannot encompass all necessary information. Addressing this issue will require the development or adoption of long-context LLMs to effectively support comprehensive evaluations.

\appendix

\onecolumn

\section{Scene Information}
We selected 10 famous novels or scripts to generate scenes. Among these, 5 are in Chinese and 5 are in English, covering different genres and themes. The details are shown in the Table~\ref{tab:sources}.
\subsection{Source for Scene Crafting}\label{source}

The following table lists the sources from which scenes are extracted for this project:
\begin{table}[h]
\centering
    \renewcommand\arraystretch{1.}
    \caption{List of sources used for scene crafting}
    \begin{threeparttable}
        \scalebox{1}{
        \begin{tabular}{ccc}
        \toprule
        Title & Type & Language \\ 
        \midrule
        Journey to the West & Novel & Chinese \\ 
        Romance of the Three Kingdoms & Novel & Chinese \\ 
        Dream of the Red Chamber & Novel & Chinese \\ 
        My Fair Princess & Novel & Chinese \\ 
        The Smiling, Proud Wanderer & Novel & Chinese \\ 
        Harry Potter & Novel & English \\ 
        The Lord of the Rings & Novel & English \\ 
        The Matrix & Script & English \\ 
        Twilight & Novel & English \\ 
        A Song of Ice and Fire & Novel & English \\ 
        \bottomrule
        \end{tabular}
        }
\end{threeparttable}\label{tab:sources}
\end{table}

\subsection{Statistics of Evaluation Scenes}\label{scene_statistics}

\begin{figure}[h]
    \centering
    \includegraphics[width=.6\linewidth]{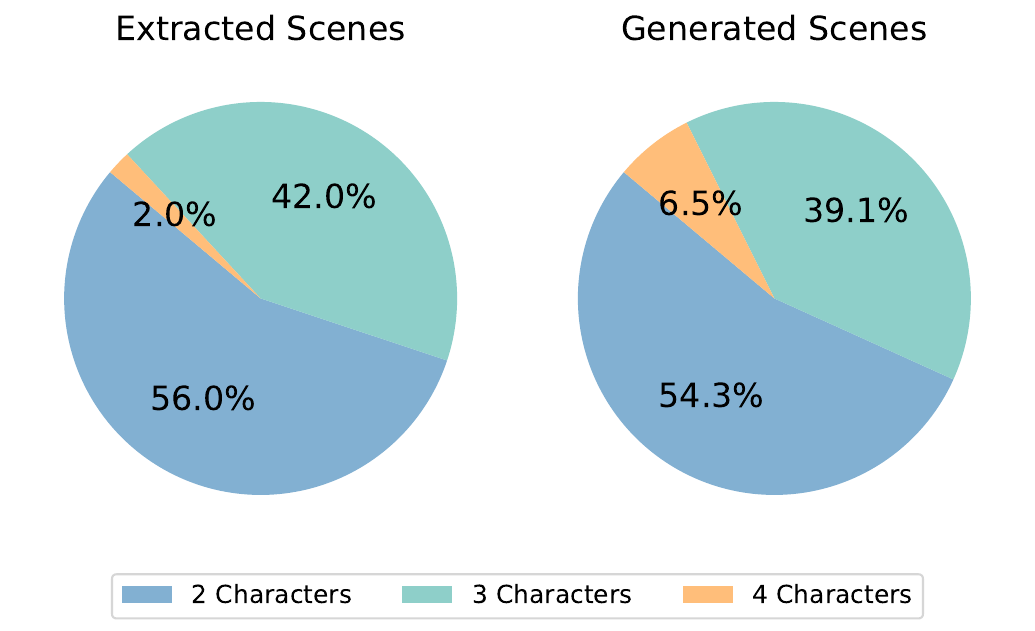}
    \caption{Distribution of characters numbers in 100 scenes.}
    \label{fig:character_100}
\end{figure}

In this section, we created five extracted scenes and five generated new scenes for each script or novel, totaling 100 scenes. The distribution of the number of characters in the 50 extracted scenes and the 50 generated scenes is shown in Fig~\ref{fig:character_100}. Most scenes feature two or three characters, with a smaller portion including four characters.

\section{Cost Analysis}

\begin{table*}[h!]

\centering
    \begin{threeparttable}
        \scalebox{1}{
\begin{tabular}{p{1.8cm}<{\centering} p{2cm}<{\centering} p{1cm}<{\centering} p{1cm}<{\centering} p{1cm}<{\centering} p{1cm}<{\centering} p{1cm}<{\centering} p{1cm}<{\centering} p{1cm}<{\centering} }
\toprule
Narrator & Character & \multicolumn{3}{p{3cm}<{\centering}}{Narrator} & \multicolumn{3}{p{3cm}<{\centering}}{Character} & Total  \\
\cmidrule(r){3-5} \cmidrule(r){6-8}
 & & Input & Output & Cost(\$) & Input & Output & Cost(\$) & Cost(\$)  \\
\midrule
GPT-3.5 & GPT-4 & 25,723 & 4,203 & 0.0192 & 75,349 & 14,407 & 0.0593 & 0.0785  \\
GPT-3.5 & GPT-3.5 & 19,954 & 3,883 & 0.0158 & 49,832 & 6,823 & 0.0352 & 0.0510  \\
GPT-3.5 & Llama-3-8B  & 24,403 & 3,928 & 0.0181 & 65,178 & 10,877 & - & 0.0181  \\
CharacterNR & Llama-3-8B  & 25,184 & 3,626 & - & 63,077 & 10,133 & - & - \\
\bottomrule
\end{tabular}
}
\end{threeparttable}
\caption{Cost for running a single scene for 3 rounds. Input is the number of tokens in the prompt fed to the LLM, Output is the number of tokens generated by the LLM, and Cost(\$) is the expense for using the OpenAI API. We selected LLama3 as a representative of open-source models. '-' indicating no external API calls or costs.}
\label{tab:cost}
\end{table*}

The evaluation of CharacterBox need supporting LLMs as narrator and evaluator. Running a single scenario for 3 rounds, the cost and time required for calling the OpenAI API are shown in the Table~\ref{tab:cost}. The local model inference is performed on a single A100 GPU. 
From the results, we can see that the main costs come from calling the GPT-3.5 API for narration and calling GPT-4 for scoring. If the number of evaluation scenarios is large, the expenses can be quite significant. Therefore, as mentioned in Section~\ref{sec:characterNR} and Section~\ref{sec:characterRM}, we fine-tuned CharacterNR and CharacterRM to serve as the narrator and evaluator, respectively, to reduce costs.

\section{Applicability to Average Character in Diverse Scene}

\begin{figure*}[t]
    \centering
    \includegraphics[width=\linewidth]{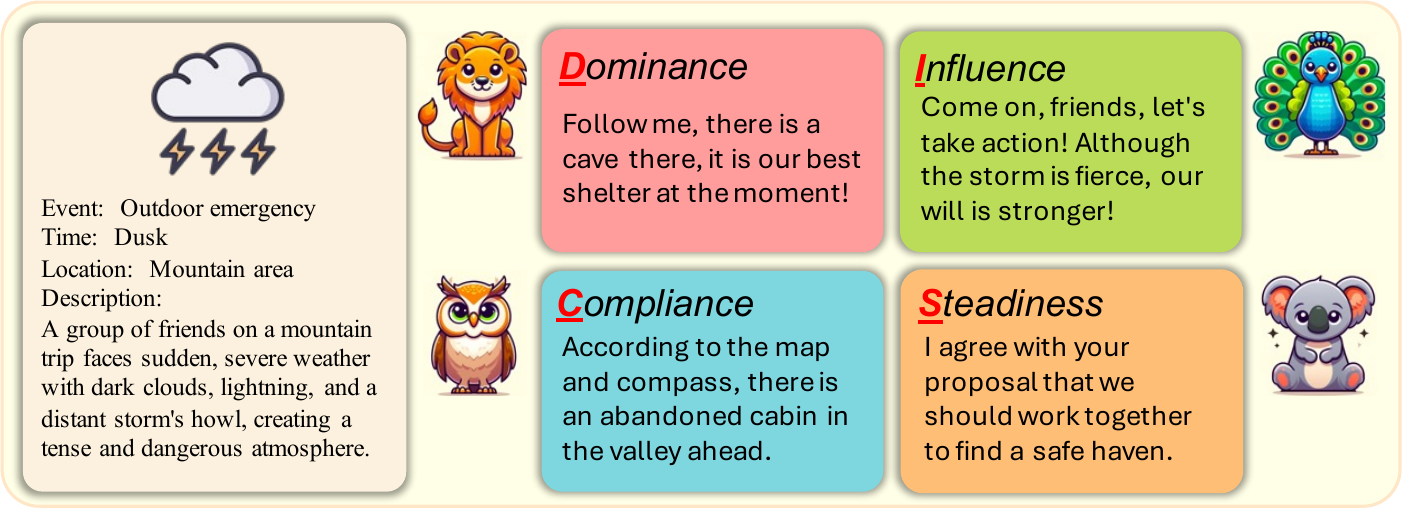}
    \caption{A case study demonstrates that CharacterBox can be extended to scenario simulations within average character in diverse contexts.}
    \label{case}
\end{figure*}

The DISC model~\cite{geier1977personal} is a psychological theory that categorizes human behavior into four types: dominance, influence, steadiness, and compliance. Dominance involves leadership and risk-taking. Influence is characterized by optimism and persuasiveness. Steadiness involves patience and supportiveness. Compliance is marked by analytical skills and precision.

To test our framework's applicability to diverse scenes, we created a challenging environment with characters of the four DISC types and observed their reactions.
As shown in Fig~\ref{case}, each character maintained their behavioral patterns in response to a sudden weather change. The dominance character led the team. The influence character boosted team confidence. The steadiness character focused on safety. The compliance character assessed risks and assisted in decision-making. This demonstrates that CharacterBox can evaluate role-playing fidelity for both famous and average characters, highlighting its potential for psychological experiments.

\section{Detailed Prompt}\label{prompt}

\subsection{Narrator Prompts}

\textbf{Action Influence}: Analyzing and describe a character's specific physical action and its tangible impact on another character.
\begin{tcolorbox}[colback=black!3!white,colframe=black!30!white,arc=0.1cm]
		Action: \textbf{[action]}\\
            Actor: \textbf{[actor]}\\
            Please analyze the physical actions and impacts detailed above, specifically focusing on the effects on ONLY one character listed in 'Characters'.\\
            Your analysis must:
            \begin{enumerate}
                \item Identify the target character affected (must be from the 'Characters' list).
                \item Describe the specific physical action initiated by the actor.
                \item Explain the tangible impact of this action on the target character's physical state or circumstances.
                \item You must pick up ONLY ONE character from the 'Characters' list.
                \item Emphasize physical interactions or impacts. If an action does not physically affect any characters listed, return the actor's name as Target Name.
                \item Must format your response as follows: [Actor];;[Target Name];;[Detailed Physical Impact of {actor} on Target].
            \end{enumerate}
            Ensure responses are concise, precise, and adhere to the specified format.
		
	\end{tcolorbox}

\textbf{Action Result}: Describing the immediate direct outcome of a character's actions concisely and clearly, focusing on the cause-and-effect relationship.
\begin{tcolorbox}[colback=black!3!white,colframe=black!30!white,arc=0.1cm]

		Action: \textbf{[action]}\\
            Instruction: Serve as an instant event adjudicator, swiftly analyzing the interactions between specified characters and their actions. Narrate the immediate outcomes in a concise omniscient narrator's voice, focusing exclusively on the direct consequences of these interactions at this very moment. Your narration should clearly and directly elucidate the cause-and-effect relationship between actions, emphasizing the instant outcomes without delving into any future implications or extended storylines.\\
            Very Important Guidelines:
            \begin{enumerate}
                \item Narrate the outcomes with immediacy, centering on the direct results of the current actions' interactions.
                \item Use a concise omniscient narrator’s voice to maintain a narrative style while ensuring the analysis is straightforward and to the point.
                \item Your analysis should be grounded in the character descriptions and actions provided, avoiding any speculative or unnecessary detail.
                \item Do not repeat the Actions in the result. The result is only the result of the current action interaction.
            \end{enumerate}
		
	\end{tcolorbox}

\textbf{Update Scene}: Making necessary adjustments solely to the physical environment based on the provided observations.
\begin{tcolorbox}[colback=black!3!white,colframe=black!30!white,arc=0.1cm]

            Given an initial scene description, examine the provided observations to identify any direct and significant physical impacts on the environment. Update the scene based on these observations, focusing solely on changes to the physical environment. If the observations do not reveal any significant physical changes to the environment, the original scene description should remain unchanged. Ensure the updated scene retains the structure of the initial scene description and does not introduce new properties that were not part of the original scene description.\\
            Note:
            \begin{enumerate}
                \item The scene description should focus solely on the physical environment and should not contain character actions or interactions.
                \item The elements 'time', 'location', and 'description' in the scene should not be changed unless the observation specifically indicates a change.
                \item The output should consist of structured elements for 'time', 'location', and 'description' without adding any extra text or prefixes.
            \end{enumerate}
            Input:
            \begin{itemize}
                \item Time: \textbf{[time]}
                \item Location: \textbf{[location]}
                \item Description: \textbf{[description]}
            \end{itemize}
            Observation: \textbf{[observation]}\\
            Output:
            \begin{itemize}
                \item Time:
                \item Location: 
                \item Description:
            \end{itemize}
		
	\end{tcolorbox}

\textbf{Update Character}: Synthesizing the character's backstory and scene observations to depict their current position and state, shaped by dynamic interactions with other characters.
\begin{tcolorbox}[colback=black!3!white,colframe=black!30!white,arc=0.1cm]
		
		Observation: \textbf{[observation]}\\
            Character Name: \textbf{[name]}\\
            Given the character's rich backstory and observation within the scene, distill this information into a succinct summary of their present location and condition.\\ 
            Focus on how their interactions, especially the dynamic interplay with other characters, shape their current circumstances.\\ 
            This interaction's effects should be evident in the nuanced portrayal of their condition and placement within the scene.\\ 
            Utilize this structured format for your depiction:\\
            Position: [Specify {name}'s exact position, incorporating environmental details or spatial context to enhance the scene's visuality.]\\
            State: [Describe {name}'s current state, weaving together emotional nuances, physical readiness, and the influence of recent encounters or developments.]
		
	\end{tcolorbox}

\subsection{Character Prompts}

\textbf{Action}: Providing a specific observable action for a character based on their personality traits and the current scene details to advance the story or character arc.
\begin{tcolorbox}[colback=black!3!white,colframe=black!30!white,arc=0.1cm]
		
        Based on \textbf{[name]}'s profile, recent memories and the current scene details, describe the next specific action \textbf{[name]} takes. This action should reflect \textbf{[name]}'s personality traits, current situation, and the physical setting. It must logically follow the scene's context and be a clear, observable act, distinct from any prior actions described.\\ 
        Avoid including dialogue or thought processes; concentrate on the physical action \textbf{[name]} is about to take. This action should be easily observable to anyone present in the scene.\\
        It is crucial that this action visibly advances the story or character arc in a way that is true to \textbf{[name]}'s character and the ongoing situation. The action should make sense within the established environment and narrative, providing a tangible progression of the scene or \textbf{[name]}'s objectives.

	\end{tcolorbox}

\textbf{Dialogue}: Crafting dialogue for a character based on their personality, observation, role in the story, and recent memory.
\begin{tcolorbox}[colback=black!3!white,colframe=black!30!white,arc=0.1cm]
		
        Based on the provided character profile and the observation, please craft a dialogue that \textbf{[name]} might say at this moment. Consider \textbf{[name]}'s personality, observation, role in the story, and the recent memory to inform the dialogue's tone and content.
		
	\end{tcolorbox}

\textbf{Reaction}: Describing a character's clear action in response to their observations, reflecting their personality, location, and state, logically fitting with what they have noticed and considering the influence of others' actions.
\begin{tcolorbox}[colback=black!3!white,colframe=black!30!white,arc=0.1cm]
		
        Based on \textbf{[name]}'s observations in the current scene, describe a clear action they take in response.\\
        This action should reflect \textbf{[name]}'s personality, location, and state, fitting logically with what they've observed, considering action influence of others actions.\\
        Focus on a visible, external action, avoiding dialogue or internal thoughts. The action must be directly related to the immediate context and observable by others.\\
        \textbf{Reminder}: The action is a response to \textbf{[name]}'s surroundings or events they've noticed.
		
	\end{tcolorbox}

\textbf{Update Self-belief}: Providing a first-person perspective on a character's self-belief, goals, and intended actions based on their current situation, observations, and recent memories.
\begin{tcolorbox}[colback=black!3!white,colframe=black!30!white,arc=0.1cm]
		
        Assuming you are now \textbf{[name]}, based on your understanding of this character, the environmental context, observation and recent memories, please describe from the first-person perspective your self-belief as this character. Focus on your identity, your current location, your state (emotional, physical, and psychological), and your goals. Reflect briefly on how this character might react, plan, and act based on their beliefs, desires, and intentions.
        \begin{enumerate}
            \item Belief: As \textbf{[name]}, what do I believe about my current situation and condition? Briefly describe your perception of yourself, highlighting key physical aspects like any injuries, your sense of movement (e.g., running, jumping), your energy levels, and any changes in physical abilities. Consider how these details influence your identity and role within the story.
            \item Desire: What are my goals? Summarize your short-term and long-term objectives, including the strategies and actions you plan to implement to achieve these goals.
            \item Intention: How do I plan to act? Outline specific actions you intend to take in pursuit of your goals, noting any potential challenges and your strategies for overcoming them.
        \end{enumerate}
        Provide concise responses shortly, focusing on your self-belief, understanding of the current situation, and future action plan.
		
	\end{tcolorbox}

\textbf{Update Env-belief}: Describing a character's belief about their environment, including perceptions of other characters, understanding of the scene, and how these factors influence their actions and decisions.
\begin{tcolorbox}[colback=black!3!white,colframe=black!30!white,arc=0.1cm]
		
	Other Characters: \textbf{[other characters]}\\
        Please act as \textbf{[name]}, given the information about other characters, the environment, and your own character's profile, please describe your belief about the environment in the first person. This includes your perception of other characters, your understanding of the scene, and how these elements influence your actions and decisions.
        \begin{enumerate}
            \item Perception of Others: Based on the interactions and information available, how do I perceive other characters? Describe your understanding of their intentions, relationships, and potential influence on your character.
            \item Understanding of the Scene: What is my understanding of the current scene and its significance to my character? Detail the environmental factors, challenges, or opportunities present.
            \item Influence on Actions: How does my perception of others and understanding of the scene influence my actions and decisions? Explain the potential strategies or reactions this insight leads to.
        \end{enumerate}
        Please provide a concise overview of your environment belief shortly, focusing on the interpersonal and environmental aspects that shape your character's perspective and future actions.
		
    \end{tcolorbox}

\section{Experimental Details} \label{trainging_detail}

The hyperparameters for training CharacterNR, CharacterRM, Guided-Qwen and Reflective-Qwen are as follows, with all models being trained using Lora and the Adam optimizer.

\begin{table*}[h]
\centering
\caption{Hyperparameter configuration for training of CharacterNR, CharacterRM, and TE-Baichuan2-7B.}
\renewcommand\arraystretch{1}
\begin{threeparttable}
    \scalebox{1}{
        \begin{tabular}{ccccc}
        \hline
        Hyperparameter & CharacterNR & CharacterRM & Guided-Qwen & Reflective-Qwen\\ \hline
        Cutoff Length & 8192 & 8192 & 8192 & 8192\\
        Per Device Train Batch Size & 1 & 1 & 1 & 1 \\
        Per Device Eval Batch Size & 1 & 1 & 1 & 1\\
        Gradient Accumulation Steps & 16 & 32 & 16 & 16 \\
        Learning Rate Scheduler Type & \texttt{cosine} & \texttt{cosine} & \texttt{cosine} & \texttt{cosine}\\
        Warmup Steps & 20 & 20 & 20 & 20\\
        Learning Rate & $5 \times 10^{-5}$ & $5 \times 10^{-5}$ & $5 \times 10^{-5}$ & $5 \times 10^{-5}$\\
        Num Train Epochs & 5.0 & 5.0 & 6.0 & 3.0\\
        Validation Size & 0.1 & 0.1 & 0.1 & 0.1\\ \hline
        \end{tabular}
    }
\end{threeparttable}\label{tab:combined_config}
\end{table*}

\section{Experiment Compute Resources} \label{resources}
The experiments conducted in this study utilized the following hardware configuration:
\begin{itemize}
    \item Operating System: Ubuntu
    \item GPU: NVIDIA 80GB A100 * 4
    \item CPU: Intel Core i7-14700KF
\end{itemize}
This setup provided the necessary computational power to efficiently handle the intensive tasks associated with our experiments, ensuring high performance and reliability throughout the study.

\section{Crowdsourcing Details} \label{annotation}
To ensure the quality and consistency of the annotation work, we invited three experts who were highly familiar with the ten selected Chinese and foreign novels, as well as the specific plots to be marked. Before starting the annotation process, the experts underwent a unified training session. Detailed reference guidelines were provided to them to standardize their work. The following sections include the full instructions given to the participants and details about their compensation.\\

\begin{tcolorbox}[colback=black!3!white,colframe=black!30!white,arc=0.1cm]
    \textbf{Mission background}\\
    Your task is to evaluate various aspects of the performance of a large language model (LLM) while performing role-playing. You will be scored on the LLM's role-playing abilities based on the following 7 indicators, with scores ranging from 1 to 5 for each indicator.
\end{tcolorbox}
    
\begin{tcolorbox}[colback=black!3!white,colframe=black!30!white,arc=0.1cm]
    \textbf{Key field descriptions}
    \begin{enumerate}
        \item Title: Which film, television or literary work the character comes from.
        \item Scene Info: Detailed information describing the background and context of the scene.
        \item Character Info: Describes the background and characteristics of the role played by the model.
        \item Behavior: The specific behavior or dialogue of the character in the scene.
        \item Knowledge Accuracy: Evaluate the accuracy of the knowledge displayed by the model in the conversation.
        \item Emotional Expression: Evaluate the way and accuracy of the model expressing emotions.
        \item Personality Traits: Evaluate the consistency and accuracy of the model in displaying the specific personality traits of the role.
        \item Behavioral Accuracy: Evaluate how accurately the model imitates and reproduces the character's behavior and language habits.
        \item Immersion: Evaluate the consistency of character performance and how it enhances user immersion.
        \item Adaptability: Assess the character's ability to adapt to new situations and changes in dialogue.
        \item Behavioral Coherence: Evaluate the logical consistency of a character's actions and responses and how they match the dialogue and plot.
    \end{enumerate}
\end{tcolorbox}

\begin{tcolorbox}[colback=black!3!white,colframe=black!30!white,arc=0.1cm]
    \textbf{Label steps}
    \begin{enumerate}
        \item Please carefully read the scene information in the [Scene Information] column and the character information in the [Character Information] column to understand the characters and their corresponding relationships.
        \item Please read the [Behavior] column carefully and use this as the main basis for scoring. The [Behavior] column records some observations (observations) and behaviors (actions) against the character's threats. Observation describes the character/observed situation that the character threatens, and Action represents the specific behavior or dialogue performed by the character in response to the current observation.
        \item Your rating should be based on the character's performance in the dialogue and how it reflects the character's knowledge, emotions, personality, behavior, consistency, cognitive and behavioral coherence.
    \end{enumerate}
\end{tcolorbox}

\begin{tcolorbox}[colback=black!3!white,colframe=black!30!white,arc=0.1cm,breakable]
    \textbf{Rating indicators}
    \begin{enumerate}
        \item Knowledge Accuracy
        \begin{itemize}
            \item 1 point: Character-related information is often wrong or irrelevant, and is clearly inconsistent with the character's background.
            \item 3 points: Information about the character is generally accurate, although occasionally there are errors or details that are not very relevant to the character's background.
            \item 5 points: Character-related information is consistently accurate and highly relevant, demonstrating in-depth knowledge and skills in the character's historical or professional background.
        \end{itemize}
        \item Emotional Expression
        \begin{itemize}
            \item 1 point: The character's emotional expression is monotonous or inappropriate, inconsistent with the dialogue content and context.
            \item 3 points: The characters' emotional expressions are moderately varied and generally match the content, but lack depth and subtlety.
            \item 5 points: The character's emotional expression is rich and profound, highly consistent with the dialogue content and context.
        \end{itemize}
        \item Personality Traits
        \begin{itemize}
            \item 1 point: The personality traits displayed often conflict with the character's setting or lack consistency.
            \item 3 points: Personality traits generally match the character's design, although there are occasional inconsistencies.
            \item 5 points: Consistently demonstrates behavior and language choices that match the character's core personality traits.
        \end{itemize}
        \item Behavioral Accuracy
        \begin{itemize}
            \item 1 point: The model fails to capture or reproduce the character’s unique behaviors and speech habits.
            \item 3 points: The model reflects the character's behavior and language habits to some extent, but is not precise or complete.
            \item 5 points: The model accurately imitates and reproduces the character's specific behaviors, language habits and mantras.
        \end{itemize}
        \item Consistency/Immersion
        \begin{itemize}
            \item 1 point: Character performance is often inconsistent, making it difficult for users to immerse themselves in or understand the character.
            \item 3 points: Character behavior is mostly consistent, but occasional contradictions slightly affect immersion.
            \item 5 points: The character's performance is consistent throughout, enhancing user immersion and effectively reflecting the character's self-awareness.
        \end{itemize}
        \item Adaptability
        \begin{itemize}
            \item 1 point: The character's performance lacks adaptability in the development of dialogue and cannot reasonably handle new situations.
            \item 3 points: The character adapts to changes in dialogue in most cases, although occasionally it may be inflexible.
            \item 5 points: The character flexibly handles any new situations in the dialogue, always maintaining character consistency and adjusting to new directions.
        \end{itemize}
        \item Behavioral Coherence
        \begin{itemize}
            \item 1 point: The characters' actions and responses are often logically confusing and do not fit the dialogue or plot development.
            \item 3 points: The character's actions and responses are generally logical and coherent, although there may occasionally be irrational aspects.
            \item 5 points: The character's actions and responses are always logically consistent and reasonably adjusted according to the dialogue and plot development.
        \end{itemize}
    \end{enumerate}
\end{tcolorbox}
\section{Broader Impact and Safeguards} \label{impact}
\textbf{Broader Impacts}~


Our proposed framework, CharacterBox, is designed to evaluate the role-playing capabilities of LLMs. It is not intended for content generation but rather for assessing the performance of LLMs in a role-playing context. The content generated by CharacterBox is contingent upon the LLMs being evaluated. Conversely, CharacterBox can be configured to assess whether the LLM generates harmful content by setting up relevant scenarios. This capability serves as a reference for the degree of alignment between the LLM's outputs and human preferences, ensuring that the LLM's behavior is guided towards ethical and socially responsible standards. By doing so, CharacterBox contributes to the broader impact of aligning technologies with human values and societal norms.

\textbf{Safeguards}~
To address the potential risks of misuse associated with CharacterBox, we have implemented stringent safeguards. These include the development of comprehensive usage guidelines that outline ethical practices and prohibit harmful content generation. 

\end{document}